\documentclass[12pt]{elsarticle}

\usepackage{graphicx} 
\usepackage{array}
\usepackage{tabularx}
\usepackage{hyperref}
\usepackage{amsfonts} 
\usepackage{breqn}
\usepackage{multirow}
\usepackage{pdflscape}
\usepackage{float}
\usepackage{caption}
\usepackage{mathrsfs}
\usepackage{amsmath}
\usepackage{setspace}
\usepackage{etoolbox}
\usepackage[labelfont=bf,labelsep=period]{caption}
\usepackage[a4paper, left=2.5cm, right=2.5cm, top=3.5cm, bottom=2.5cm]{geometry}

\biboptions{sort&compress}
\hypersetup{
 colorlinks,
 citecolor=black,
 filecolor=black,
 linkcolor=black,
 urlcolor=black
}

\begin{document}
\begin{frontmatter}

\title{\textbf{Real-Time Surrogate Modeling for Personalized Blood Flow Prediction and Hemodynamic Analysis}}

\affiliation[aff1]{organization={Laboratory of Hemodynamics and Cardiovascular Technology, EPFL},
 city={Lausanne},
 country={Switzerland}}

\affiliation[aff2]{organization={Stanford DBDS and HAI, Stanford University},
city={Palo Alto},
country={USA}}

\affiliation[aff3]{organization={Medical School, National and Kapodistrian University of Athens},
 city={Athens},
 country={Greece}}

\author[aff1]{Sokratis J. Anagnostopoulos}
\author[aff1]{George Rovas}
\author[aff2]{Vasiliki Bikia}
\author[aff3]{Theodore G. Papaioannou}
\author[aff3]{Athanase D. Protogerou}
\author[aff1]{Nikolaos Stergiopulos}
 
\begin{abstract}
Cardiovascular modeling has rapidly advanced over the past few decades due to the rising needs for health tracking and early detection of cardiovascular diseases. While 1-D arterial models offer an attractive compromise between computational efficiency and solution fidelity, their application on large populations or for generating large \emph{in silico} cohorts remains challenging. Certain hemodynamic parameters like the terminal resistance/compliance, are difficult to clinically estimate and often yield non-physiological hemodynamics when sampled naively, resulting in large portions of simulated datasets to be discarded. In this work, we present a systematic framework for training machine learning (ML) models, capable of instantaneous hemodynamic prediction and parameter estimation. We initially start with generating a parametric virtual cohort of patients which is based on the multivariate correlations observed in the large Asklepios clinical dataset, ensuring that physiological parameter distributions are respected. We then train a deep neural surrogate model, able to predict patient-specific arterial pressure and cardiac output ($CO$), enabling rapid a~priori screening of input parameters. This allows for immediate rejection of non-physiological combinations and drastically reduces the cost of targeted synthetic dataset generation (e.g. hypertensive groups). The model also provides a principled means of sampling the terminal resistance to minimize the uncertainties of unmeasurable parameters. Moreover, by assessing the model's predictive performance we determine the theoretical information which suffices for solving the inverse problem of estimating the $CO$. Finally, we apply the surrogate on a clinical dataset for the estimation of central aortic hemodynamics i.e. the $CO$ and aortic systolic blood pressure ($cSBP$).
\end{abstract}

\end{frontmatter}

\newpage

\section{Introduction}

Physics-based modeling of arterial hemodynamics is a powerful tool for noninvasive assessment of cardiovascular function, because pressure and flow waveforms contain information about clinically relevant quantities like the cardiac output ($CO$), arterial stiffness, and peripheral vascular resistance. Among available modeling approaches, 1-D pulse-wave models that perform a cross-sectional average solution of the  the incompressible Navier-Stokes equations coupled with a wall compliance equation constitutes a very effective compromise: They can capture wave propagation, reflection, and dispersion throughout large arterial networks at a cost that is orders of magnitude lower than 3-D CFD modeling, while retaining clinically relevant data (e.g., central and peripheral pressures, pulse wave velocity, and flow distribution) \cite{shi2011review,formaggia2003oned,sherwin2003vascular}. These models are mature validated tools for blood flow prediction, allowing for systematic comparisons with in vitro replicas and in vivo measurements \cite{matthys2007pulse,reymond2009validation}, and they form a standard basis for studying how anatomy, stiffness, and distal loading affect the pressure-flow response characteristic. \cite{alastruey2009pattern}.

A notable extension of our 1-D model \cite{reymond2009validation} has been the personalization of the parameters of the 1-D model based on readily available noninvasive measurements, and the subsequent estimation of aortic cardiovascular properties of interest in an inverse solution approach. The devised methods combine multiple forward optimization loops to calibrate the 1-D model and render it patient-specific before deriving the desired central hemodynamic indices. The performance of the inverse methods is assessed via comparison of the estimated central quantities with their reference (ground truth) values, which are either experimentally measured (typically invasively for \emph{in vivo} datasets) or readily available (\emph{in silico} datasets). Another major clinical indicator of cardiac (left ventricular) function is the cardiac output ($CO$), defined as the volume of blood pumped by the heart into the systemic circulation, expressed in liters per minute. In an inverse problem-solving manner, patient-specific hemodynamics at the aortic root ($CO$ and central systolic pressure) were accurately derived from noninvasive cuff-pressure and pulse wave velocity ($PWV$) measurements in a small set of healthy adults \cite{bikia2019noninvasive}.

An important target in both modeling and clinical translation is arterial stiffness, which is commonly expressed by the carotid-femoral pulse wave velocity ($cfPWV$). The latter has proven to be a robust, prognostically important indicator that can provide standardization guidance on distance measurements and results interpretation \cite{laurent2006stiffness,vanbortel2012cfpwv}. $cfPWV$ can therefore be used to calibrate or scale the arterial network compliance in a physiologically interpretable manner, thus allowing for individual-specific variation without the need of invasive data. However, even with a calibrated one-dimensional model, large-scale tasks such as uncertainty quantification, sensitivity mapping in multidimensional parameter spaces, and inverse inference (e.g. estimation central hemodynamics from peripheral pressures) can be computationally prohibitive when they require many forward solutions.

This motivates the development of surrogate models for rapid simulation of the one-dimensional solver, which, while preserving the parameter-response relationship, can produce pressures, flows, and derived indices in near real time. Recent work has demonstrated automated steps for creating and simulating reduced-order (0D/1D) cardiovascular models from anatomical data \cite{pfaller2022automated}, as well as data-driven reduced-order simulators that learn the dynamic behavior of a system (e.g., through graph-based representations) while maintaining accuracy over a variety of geometries and boundary conditions \cite{pegolotti2024gnn}. Machine learning (ML) techniques have recently been explored, in order to exploit the information of available virtual and clinical datasets. Bikia et al. \cite{bikia2021determination} utilized a synthetic database generated by the 1-D cardiovascular model and used regression analysis to simulate various hemodynamic states. A novel ML method was also introduced to estimate the total arterial compliance via exploiting the information from a carotid blood pressure waveform, as well as typical human data from an available broad longitudinal study for the development and progression of cardiovascular disease \cite{bikia2021assessment} and cardiac contractility and $CO$ \cite{bikia2020noninvasive}. However, constructing appropriate \emph{in-silico} datasets whose parameters represent realistic cases is crucial to avoid training surrogates on unlikely combinations. Large, well-characterized population studies provide a valuable basis for such well correlated sampling \cite{rietzschel2007asklepios}, while space-filling designs, such as Latin Hypercube Sampling (LHS), allow for controlled extrapolation to effectively cover extreme parameter values \cite{mckay1979lhs}.

While our 1-D in-house model has proven to be an invaluable and adaptable tool for various subsequent investigations, the optimization process of inversely solving patient-specific cases is tedious and requires multiple forward evaluations (corresponding to hours of computational time). Moreover, the generation of specialized \emph{in silico} datasets is highly time consuming and produces a large proportion of non physiological conditions when the hemodynamic parameters are randomly sampled. To address these limitations, we develop an ML surrogate model for real-time hemodynamic parameter inference to study the generalized cardiovascular problem. The contributions of this work are the following:

\begin{itemize}
    \item Generation of parametric \emph{\emph{in silico}} cohort which reproduces the multivariate correlations of the hemodynamic parameters in Asklepios clinical dataset, ensuring physiological statistical fidelity.
    \item Design and training of ML surrogate model for real-time prediction of generalized hemodynamics and a~priori validation of candidate parameter sets, allowing non-physiological cases to be rejected prior to simulating. This accelerates the generation of synthetic datasets and provides a better strategy for sampling terminal resistance and compliance, which are often unknown.
    \item Deployment of the surrogate model for pairwise sensitivity analysis across the hemodynamic parameter space, revealing dominant interactions and their effect on arterial pressure. Production of generalized response surfaces for systolic and diastolic pressure as functions of the most influential parameters, providing an interpretable map of pressure dynamics.
    \item Investigation of non-uniqueness of the inverse patient-specific solutions for $CO$, by quantifying the training performance while different input parameter sets are used, revealing the significance of either the terminal compliance/resistance. Application to clinical dataset for noninvasive prediction of $CO$ and $cSBP$ from cuff measurements.
\end{itemize}

In this context, the present work combines a validated 1-D arterial network surrogate with physiologically that operates in both forward and inverse modes. The result is a robust framework that reproduces the network hemodynamics in an acceptable physiological domain, allows for rapid sensitivity analysis and parameter value mapping, and supports clinically generated inverse queries of central hemodynamic quantities using noninvasive measurements.

\section{Methods}

\subsection{1-D arterial model}

Arteries are typically viewed as elongated conical segments with a viscoelastic wall. The 1-D continuity and momentum equations (Navier-Stokes) are solved iteratively for an arterial network consisting of 103 segments, along with a constitutive law for the flexible arterial walls \cite{reymond2009validation}, described by the following closed-system of PDEs:
\begin{equation}
\frac{\partial Q}{\partial t} + \frac{\partial}{\partial x} \left( \int_A u^2dA\right) = -\frac{A}{\rho}\frac{\partial P}{\partial x} - 2\pi R \frac{\mu}{\rho} \frac{\partial u}{\partial r} \Bigg|_{r=R} + A f_x,
\label{p1:eq1}
\end{equation}
\begin{equation}
\frac{\partial P}{\partial t} = -\frac{1}{C_A}\left(\frac{\partial A^v}{\partial t} + \frac{\partial Q}{\partial x}\right),
\label{p1:eq2}
\end{equation}
\begin{equation}
A(t) = A^e(P(t)) + A^v(t),
\label{p1:const}
\end{equation}

\noindent where $A(x, t)$ is the arterial lumen area of radius $R(x, t)$, $Q(x, t)$ is the volumetric blood flow rate, $P(x, t)$ is the transmural pressure, $\mu$ is the blood viscocity, $f_x$ is the gravitational force and $u$ is the velocity profile derived by Womersley theory. The constitutive law (Eq. \ref{p1:const}) depends on the non-linear elastic $A^e$ and viscoelastic $A^v$ components, which are a function of the pressure and area compliance $C_A$. At the terminal arterial sites, a 3-element Windkessel model is used to simulate the omitted branches 

\begin{equation}
\frac{\partial Q}{\partial t} = \frac{1}{R_1}\frac{\partial P}{\partial t} + \frac{P}{R_1R_2C_T} - \left(1+\frac{R_1}{R_2}\right)\frac{Q}{R_1C_T},
\label{p1:wk}
\end{equation}

\noindent where $R_T=R_1+R_2$ is the terminal resistance, and $C_T$ is the terminal compliance parameters, which characterize the resistance and capacitance properties of the omitted capillaries and can be fine-tuned based on the patient.

\subsection{Dataset generation}

To develop a generalized surrogate model of cardiovascular flow, the 1-D arterial model was deployed to generate an \emph{in silico} dataset of 2000 patients, covering a broad range of patient hemodynamic properties. The input features were the $CO$, the heart rate ($HR$) and several global multipliers for the arterial length ($\lambda_L$), diameter ($\lambda_D$), terminal resistance ($\lambda_{R_T}$), arterial compliance (proportional to distensibility) ($\lambda_{C}$) and terminal compliance ($\lambda_{C_T}$). To generate a dataset with realistic parameter correlations, instead of randomly sampling the patient-specific parameters, we used the Asklepios dataset to draw physiologically correlated values within the parameter space which can be estimated noninvasively ($L, D, CO, DBP, SBP, C$). For the Windkessel coefficients $R_T$ and $C_T$, since experimental measurement is not feasible in clinical practice, we chose to perform uniform random sampling such that the range of the total arterial resistance and compliance were within [0.10, 3.80] mL/mmHg and [0.40, 2.00] mmHg$\cdot$s/mL, respectively \cite{langewouters1982visco,segers2008three,lu2006continuous}. This random sampling approach only for the Windkessel parameters was adopted to provide the full range of possible combinations such that their effect on the rest of the patient hemodynamic parameters can be thoroughly studied. All the global multipliers scale a reference arterial network \cite{reymond2009validation}, with the arterial length ($L$) scaled with the patient height and the arterial diameter ($D$) scaled with the body surface area (BSA), the age and the gender, following the literature \cite{wolak2008aortic}, to match each new patient characteristics. The carotid-to-femoral pulse wave velocity ($cfPWV$) is used to derive the arterial compliance (or distensibility) multiplier of the arterial network. More specifically, we calculate the arterial distensibility multiplier $\lambda_C$ which fits the $cfPWV$ measurement, by considering the individual arterial compliancies of all segments that lie in between the carotid and the femoral arteries as:

\begin{equation}
\text{cfPWV} = \frac{\sum_i L_i}{\sum_i L_i \sqrt{\rho \lambda_C \cdot D_i}},
\end{equation}

The clinical dataset from round 1 of the longitudinal population study Asklepios, consisting of n=2524 (1301 women) participants aged between 35-55 years, free from apparent cardiovascular diseases \cite{AsklepiosSegers2007,AsklepiosRietzschel2007,rietzschel2007rationale}, is composed of measurements for the stroke volume ($SV$) (volume of blood pumped out of the heart's left ventricle over one heartbeat), which is measured using Doppler imaging. Multiplied by the heart rate, $SV$ yields the $CO$ value. Moreover, brachial cuff pressures are measured for all patients, as well as the $cfPWV$ values.

The selected ranges of the \textit{in silico} cohort for each of the parameters are shown in Table \ref{p1:t1}. For this study we chose to omit the outliers present in Asklepios study which had values extending outside of $3\sigma$ of the parameter distributions. Hence, the final subset of the Asklepios dataset had a total of 1500 patients. To further enhance the dataset, quantile Latin Hypercube Sampling (LHS) was applied to inject 500 additional virtual patients that followed similar correlations as the Asklepios subset, such that a total of 2000 patients were finally simulated by the 1-D code, ensuring that each variable is represented across its entire range. We note that this method could be applied for generating an arbitrary number of virtual patients with realistic parameter correlations to augment the initial dataset.

\begin{table}[H]
\centering
\caption[\emph{In silico} dataset parameter range and distribution.]{\emph{In silico} dataset parameter range and distribution.}

\begin{tabular}{cccc}
\hline
\textbf{Parameter} & \textbf{Range} & \textbf{Mean} & \textbf{SD} \\
\hline
Age (years) & 35 - 57 & 45.76 & 5.61 \\
PWV (m/s) & 4.5 - 9.0 & 6.4 & 0.92 \\
Height (cm) & 152 - 190 & 169.12 & 7.92 \\
Weight (kg) & 48 - 105 & 72.89 & 12.13 \\
SBP (mmHg) & 102 - 169 & 129.64 & 14.08 \\
HR (bpm) & 43 - 85 & 63.1 & 8.07 \\
CO (L/min) & 2.8 - 7.5 & 5.1 & 0.96 \\
\hline
\label{p1:t1}
\end{tabular}
\end{table}
\vspace{-1em}
\subsection{Neural network surrogate}
We consider a fully connected neural network with two core modes. The forward mode is deployed to map seven hemodynamic features to the diastolic and systolic blood pressures of the brachial artery:
\begin{equation}
f_\theta(\lambda_D,\lambda_L, \lambda_C, \lambda_{C_T}, \lambda_{R_T}, HR,CO,)
= (DBP,\ SBP),
\qquad
f_\theta:\ \mathbb{R}^7 \longrightarrow \mathbb{R}^2.
\end{equation}
For the inverse mode, we take advantage of only the parameters that can be estimated in clinical practice to provide information about the central hemodynamics, namely the $CO$ and the central systolic blood pressure ($cSBP$) inside the ascending aorta:
\begin{equation}
f_\theta(\lambda_D,\lambda_L, \lambda_C, \lambda_{C_T}, \lambda_{R_T}, HR,DBP,SBP)
= (CO,\ cSBP),
\qquad
f_\theta:\ \mathbb{R}^8 \longrightarrow \mathbb{R}^2.
\label{p1:inv_model}
\end{equation}

\begin{figure}[H]
 \centering
 \includegraphics[width=1.0\textwidth]{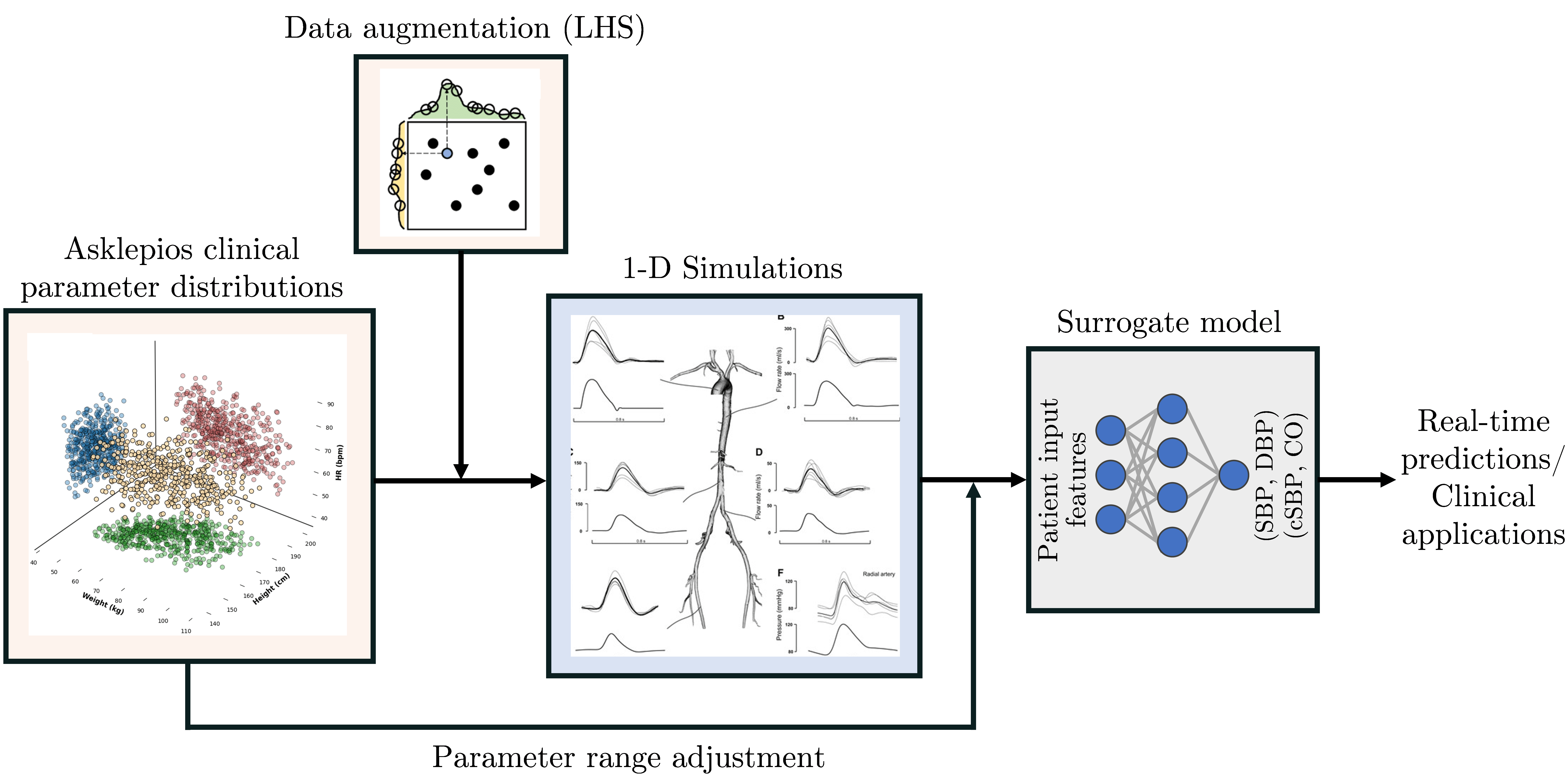}
 \caption[Real-time prediction workflow]{\textbf{Real-time prediction workflow}: The Asklepios dataset is used to draw physiologically correlated values within the parameter space which can be clinically estimated noninvasively. The dataset is augmented using LHS method and 2000 1-D forward simulations are performed using the 1-D cardiovascular code. The parameter space can be adjusted depending on the physiological range depending on the study. The surrogate model is trained either in forward or inverse mode to provide real-time hemodynamic predictions.}
 \label{p1:workflow}
\end{figure}

The schematic representation of the complete workflow is depicted in Fig. \ref{p1:workflow}.

The ML model chosen for this study is a fully connected neural network designed to map the hemodynamic features to the output features ($DBP, SBP$) for forward or ($CO, cSBP$) for inverse. Its architecture follows a symmetric expansion-compression pattern with hidden layers of 128, 256, and 128 neurons with batch-normalization, each using the $\tanh$ activation and a dropout rate of 0.1 to reduce variance. A linear output layer yields the target predictions. All inputs and outputs are standardized to zero mean and unit variance, which stabilizes the optimization landscape and improves gradient flow. The smooth L1 loss (Huber loss) is chosen to provide quadratic sensitivity for small residuals while remaining robust to occasional outliers in blood-pressure measurements. Its piecewise form behaves as an $\ell_2$ loss for $|y - \hat{y}| < \beta$ and transitions to a linear penalty beyond that threshold, preventing gradient explosions and improving stability when the data include measurement noise or imperfect labels, common in physiological sensing tasks.

\begin{equation}
L_{\delta}(y, \hat{y}) =
\begin{cases}
\frac{1}{2}(y - \hat{y})^{2}, & \text{if } |y - \hat{y}| < \beta,\\[6pt]
\beta\left(|y - \hat{y}| - \tfrac{1}{2}\beta\right), & \text{otherwise}.
\end{cases}
\label{p1:l1}
\end{equation}

\begin{figure}[H]
 \centering
 \includegraphics[width=0.95\textwidth]{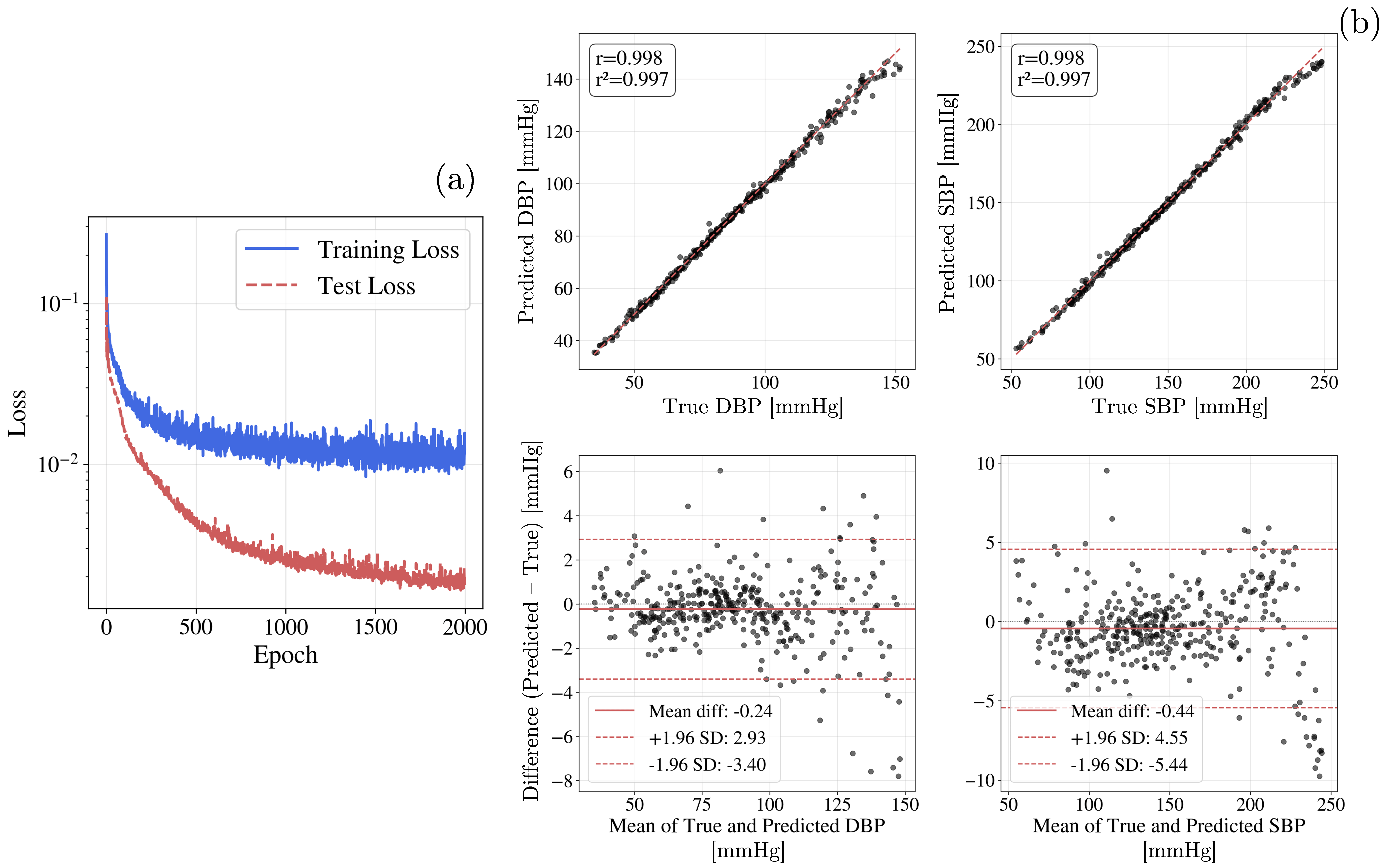}
 \caption[Training performance]{\textbf{Training performance:} Indicative convergence of the forward surrogate mode. (a) Mean-squared error loss over 2000 epochs for training and test sets. The convergence minimum indicates no overfitting and good model generalization. (b) Correlation and Bland-Altman plots: agreement between predicted and measured blood pressures for the brachial DBP (left) and SBP (right) network outputs.}
 \label{p1:loss}
\end{figure}

Training is performed using AdamW with a learning rate of $3\times 10^{-4}$, a weight decay of $10^{-4}$, batch size 128, and an 80/20 train-test split under a fixed random seed for reproducibility. Indicative convergence curves are shown in Fig.~\ref{p1:loss}(a) for the forward mode, which show consistent monotonic reduction of both training and test MSE, with the test loss descending more sharply due to stronger regularization effects from dropout at inference time. The absence of divergence or widening separation between curves indicates no signs of overfitting, and the late-epoch plateau demonstrates that the optimizer approaches a stable minimum. The correlation and Bland-Altman plots in Fig.~\ref{p1:loss}(b) further confirm reliable predictive behaviour: the regression outputs exhibit strong linear agreement with ground truth and narrow limits of agreement, with no systematic bias across the pressure range. Excluding a few outliers at the extrema, the $DBP$ and $SBP$ errors lie within ± 5 mmHg. Together, the convergence trends and error distributions indicate a well-behaved model with robust generalization properties.

\section{Results}
\subsection{Sensitivity analysis}

\begin{figure}[H]
 \centering
 \includegraphics[width=1.0\textwidth]{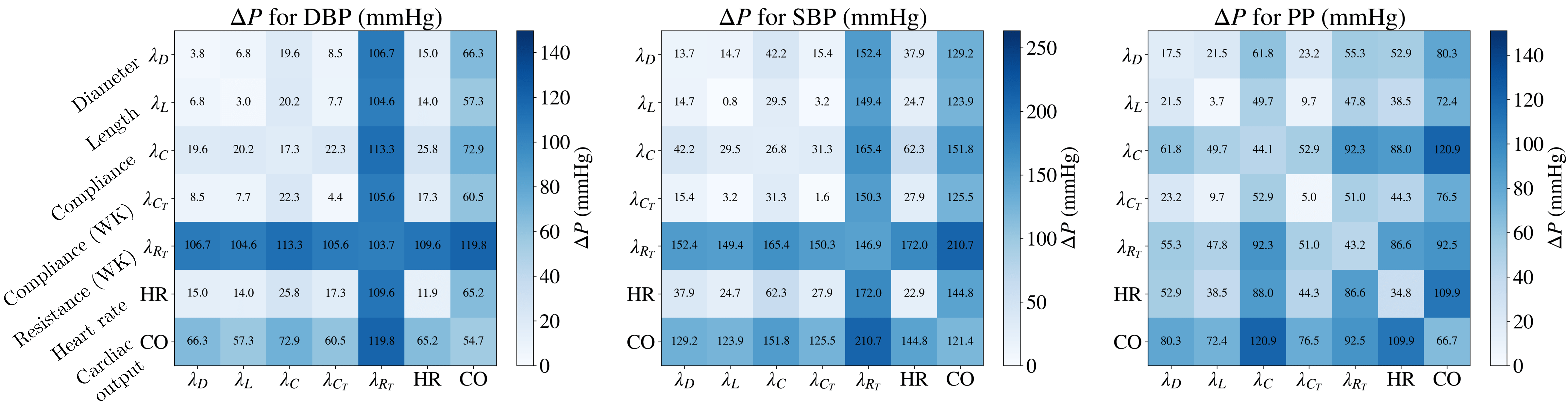}
 \caption[Pair-wise parameter sensitivity]{ \textbf{Pair-wise parameter sensitivity:} The sensitivity heatmap of varying pair-wise parameters (while the rest are held fixed), reveals the parameters with the highest influence on the diastolic/systolic blood pressure are the terminal resistance $R_T$, the $CO$, the compliance $C$ and the $HR$. For the pulse pressure $(PP)$, the most dominant parameters are the $CO$, the $R_T$ and $R_C$. The geometry, as well as the terminal compliance have negligible influence in the $\Delta P$ over the full parameter range variation.}
 \label{p1:delta}
\end{figure}

\begin{figure}[H]
 \centering
 \includegraphics[width=1.0\textwidth]{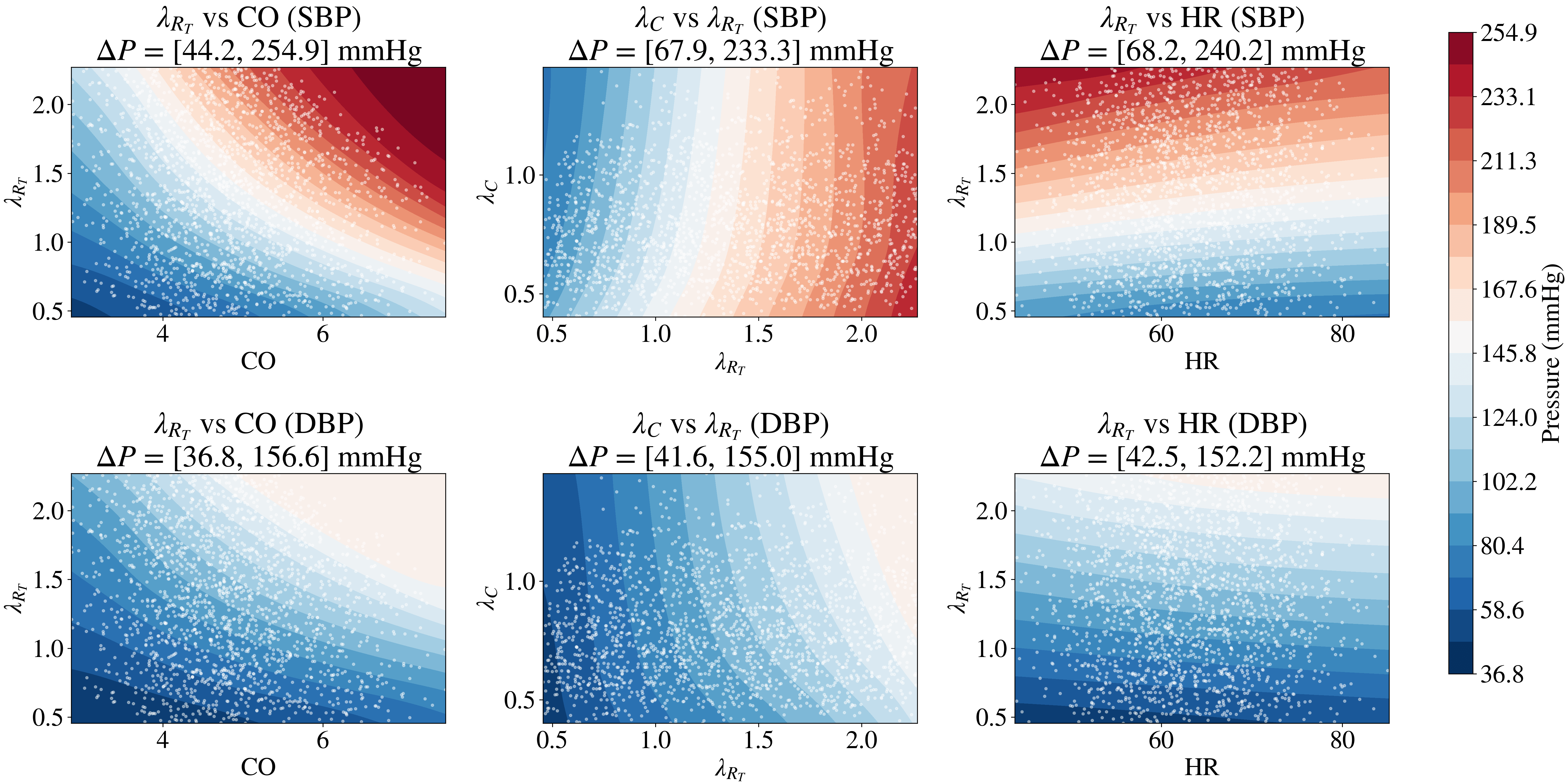}
 \caption[Pair-wise contour maps]{ \textbf{Pair-wise contour maps:} The hemodynamic interactions between each parameter pair (while the rest are held fixed) cause distinct 2-D variations in the resulting pressure fields for both $SBP$ (top), and $DBP$ (bottom). The white points represent the \textit{in silico} patients used for the training.}
 \label{p1:sbp}
\end{figure}

The primary criterion for accepting a virtual subject as physiological forward solution is the requirement that the predicted $SBP$ and $DBP$ fall within their admissible ranges. To quantify which parameters have the strongest influence on arterial pressure, we construct three square pairwise matrices (Fig. \ref{p1:delta}) in which each entry 
represents the pressure variation across the full range of a given parameter pair, while the remaining parameters are kept fixed at a reference value. The diagonal shows the sensitivity of each individual parameter. Fig. \ref{p1:sbp} illustrates the six most sensitive pairs for representative comparisons. All continuous fields are generated in near real time by the trained DNN, enabling exhaustive exploration of the parameter space.

From the sensitivity matrix of Fig. \ref{p1:delta} we see that the geometric scaling parameters $\lambda_D$ and $\lambda_L$ produce only minor changes in $\Delta P$ of $DBP$ and $SBP$, reflecting their limited hemodynamic influence. Therefore we can safely assume that simply scaling a reference geometry can accurately model most of the healthy arterial networks (e.g. without aneurysms or stenoses). Likewise, the terminal compliance $C_T$ exhibits weak influence, consistent with its contribution of only a small fraction of the total arterial compliance. In contrast, $R_T$, $CO$, $C$ and $HR$ emerge as the dominant factors shaping the pressure response. For the pulse pressure $(PP)$ (given by SBP-DBP), the most dominant parameters are the $CO$, the $R_T$ and $R_C$. The pairwise contour maps between the more influencing parameters are plotted in Fig. \ref{p1:sbp} for $SBP$ (top) and $DBP$ (bottom), overlaid with the training patient dataset (in white), provide a clear depiction of how hemodynamic interactions cause the resulting pressure ranges, and reveal the correlation patterns that govern the system’s global behavior.

\subsection{Arterial pressure study}

\begin{figure}[H]
 \centering
 \includegraphics[width=0.9\textwidth]{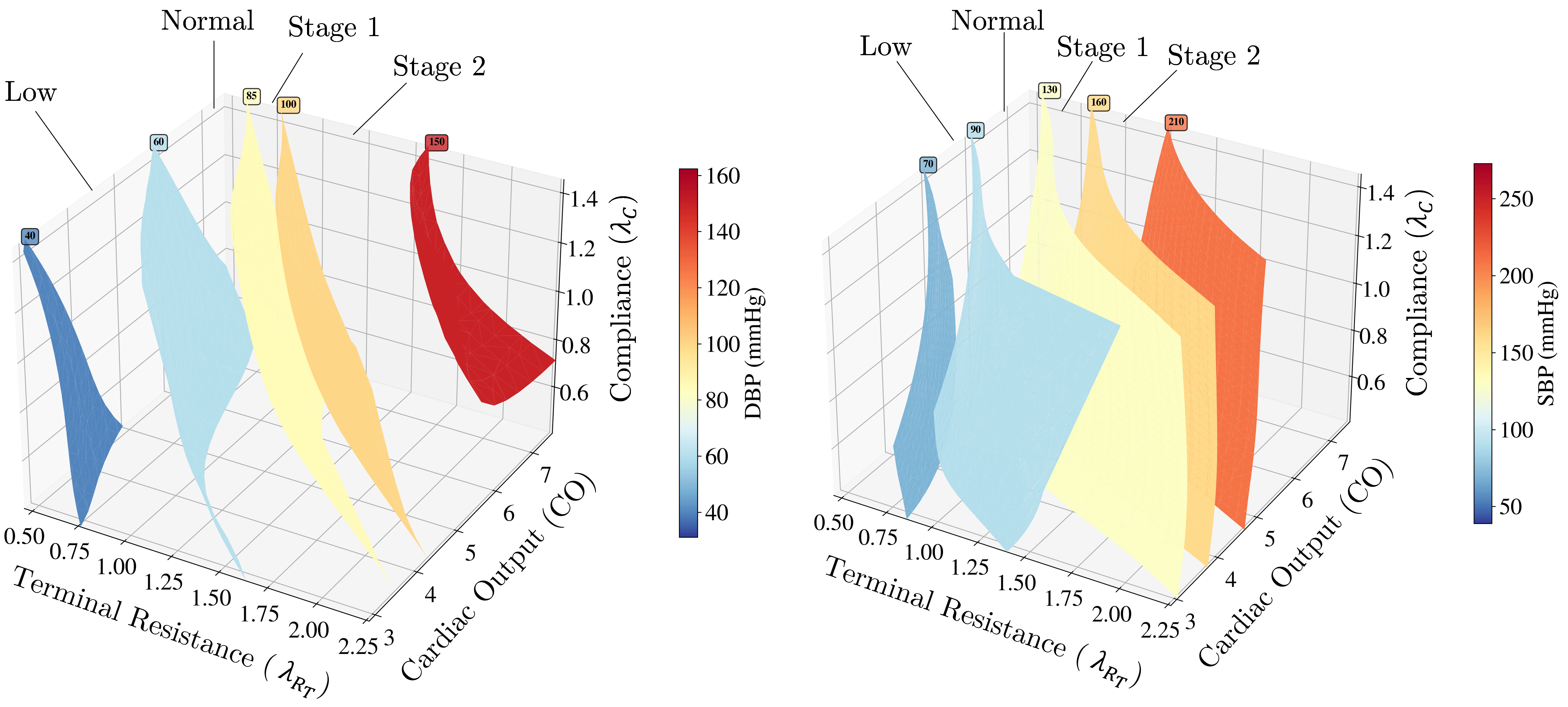}
 \caption[Pressure iso-surfaces]{ \textbf{Pressure iso-surfaces:} Plotting the most sensitive hemodynamic parameters reveal distinct regions of low, normal and hypertensive patients for the $DBP$ (left) and $SBP$ (right). The regions outside this range are non-physiological cases. }
 \label{p1:4d}
\end{figure}

To better visualize the physiological domains associated with the most sensitive input 
parameters, Fig. \ref{p1:4d} displays 4-D iso-surfaces illustrating the 
regions corresponding to low, normal, stage~1, and stage~2 hypertensive groups. 
Points lying outside the extremal pressure bounds are classified as 
non-physiological. These surfaces reveal the global structure of the pressure fields 
within the cardiovascular network as $CO$, $R_T$, and $C$ are varied 
simultaneously, exposing how changes in flow, peripheral resistance scaling, and 
arterial compliance jointly shape the admissible hemodynamic space.

It is worth noting that these iso-surfaces were constructed from a 
$30\times30\times30$ grid of DNN predictions evaluated in real-time. Comparable 
manifolds can be generated for any other combination of input and output variables, 
enabling systematic exploration of the multidimensional parameter-response landscape. 
For context, computing this many forward simulations with the original physics-based 
model would require more than $\sim\!50$ CPU days on a single core, which emphasizes 
the computational flexibility provided by the surrogate model.

\subsection{Real population matching}
\label{p1:matching}

So far, we have shown that the surrogate outperforms the full solver in predicting generalized fields of virtual patients in near-real time. We now examine whether constraining the pressure range yields realistic distributions comparable to a large clinical dataset such as Asklepios. The distributions of all known input parameters ($D$, $L$, $cfPWV$, $HR$, $CO$) have already been matched, so together with the injected LHS points they closely follow the clinical statistics. The two unknown parameters ($R_T$, $C_T$), however, were sampled randomly and may strongly influence the numerical output. 

\begin{figure}[H]
 \centering
 \includegraphics[width=1.0\textwidth]{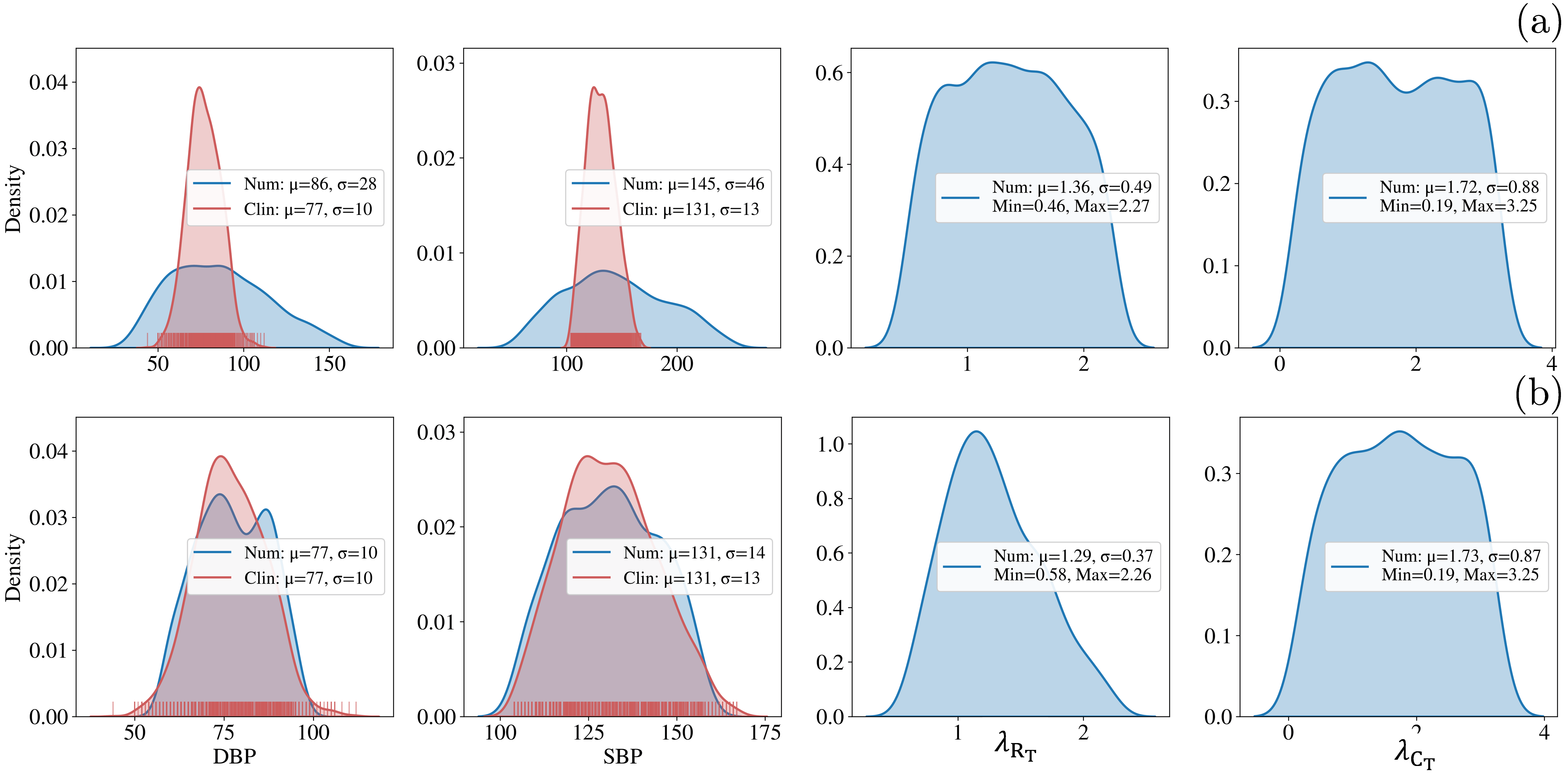}
 \caption[Population matching]{ \textbf{Population matching:} (a) The distribution of naively sampled terminal resistance ($R_T$) widens the output pressure distributions, even though the rest of the known parameters are sampled directly from the clinical dataset of Asklepios. (b) Constraining only the range of pressures indicates that the 1-D model is able to closely match the clinical distributions of Asklepios for both $DBP$ and $SBP$, while the distribution of $R_T$ takes a triangular shape.} 
 \label{p1:p_matching}
\end{figure}

Our aim is therefore to assess whether constraining only the numerical outputs of diastolic and systolic pressures ($DBP$, $SBP$), is sufficient to bring their distributions in line with the clinical ones. To this end, we extract 620 patients from the full set of 2000 whose pressures fall within $95\%$ of the corresponding Asklepios ranges (i.e., within $\pm 1.96$ SD). As shown in Fig. \ref{p1:p_matching}, the numerical model initially produces broad pressure distributions despite sampling the known parameters from the clinical dataset. Once we enforce only the pressure ranges, however, the shapes of the $DBP$ and $SBP$ distributions align closely with their clinical counterparts.

Interestingly, the range of $R_T$ remains almost unchanged (consistent with values reported in the literature), but its initially uniform sampling shifts toward a triangular distribution. In contrast, the distribution of $C_T$ remains essentially unaffected, reflecting its low sensitivity in this setting. These findings both support the validity of the 1-D modeling framework and indicate that, for population matching, the dominant factor is the sampling distribution of the terminal resistance, while its admissible range is already well established in the literature. Finally, we can easily train a custom neural model to produce the mapping $f_\theta(D, L, cfPWV, HR, CO, SBP, DBP) \rightarrow R_T$, yielding a Pearson coefficient of 0.998 (not shown here), that reproduces exactly the numerical distributions of $DBP, SBP$ and $R_T$, shown in Fig. \ref{p1:p_matching} (b). This mapping enables the generation of any \emph{in silico} dataset composed solely of admissible patients that match any given \emph{in vivo} population, given that the $CO$ distribution is also available as input.


\subsection{Inverse solution uniqueness}

\begin{figure}[H]
 \centering
 \includegraphics[width=1.0\textwidth]{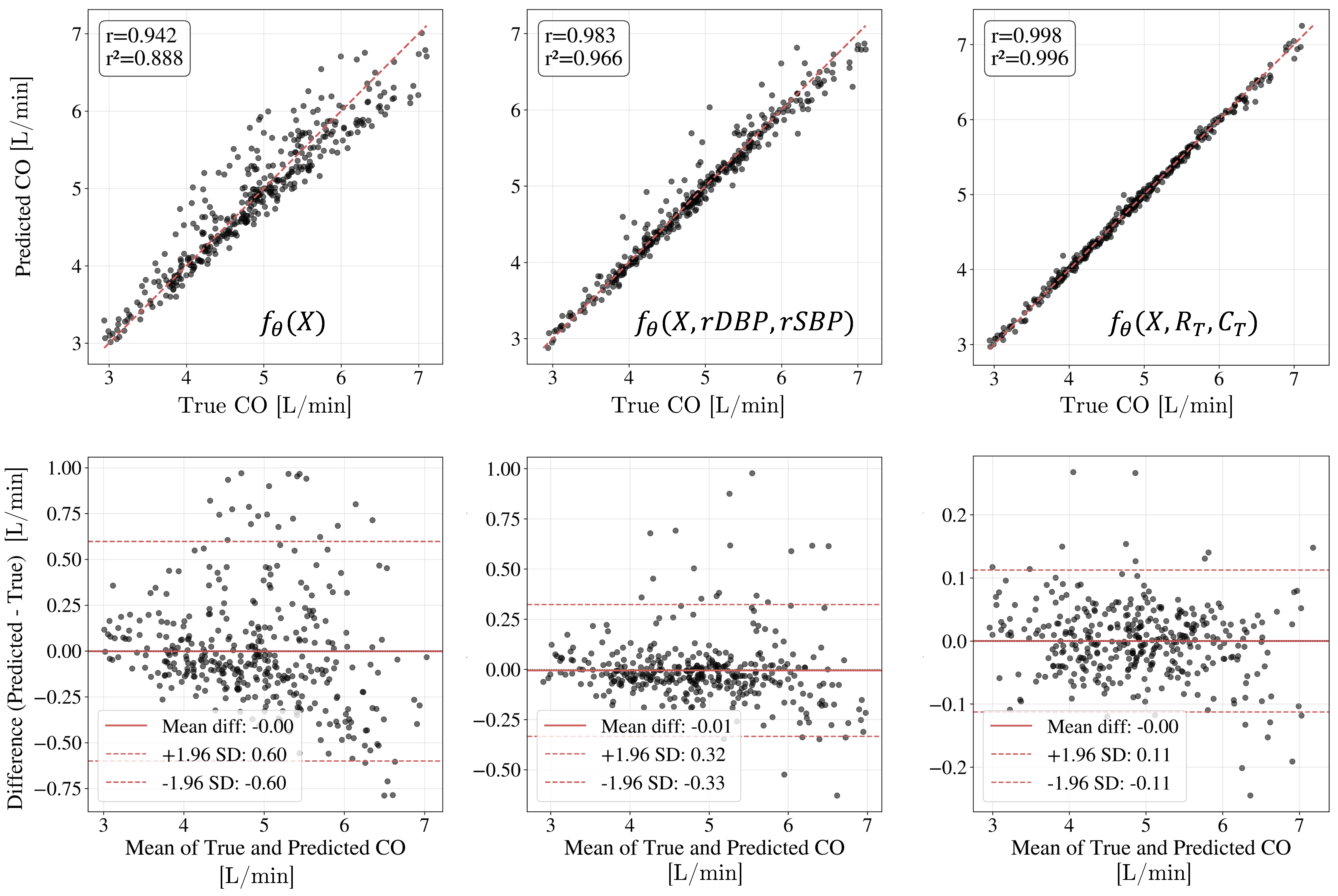}
 \caption[$CO$ identification]{ \textbf{$CO$ identification:} (a) Using only the noninvasively obtainable parameters ($L$, $D$, $C$, $HR$, $SBP$, $DBP$) yields high correlation but large limits of agreement, reflecting a strong indication of non-uniqueness for the inverse mapping. (b) Introducing an extra pressure reading at the radial artery (b) or the terminal resistance and compliance ($R_T$, $C_T$) (c), resolves the ambiguity and recovers $CO$ with near-perfect accuracy.}
 \label{p1:2by4}
\end{figure}

To demonstrate the versatility of the surrogate, we deploy different versions of the model of Eq. \ref{p1:inv_model} to perform a study on the inverse estimation of the $CO$ on the matched virtual population of the previous section, considering three main inverse scenarios: (a) given only the parameters which can be clinically estimated (no information about Windkessel parameters ($R_T, C_T$), (b) introducing an additional pressure measurement at the radial artery ($rSBP, rDBP$) which complements the $(SBP, DBP)$ measurements at the brachial artery, and (c) by using the full parameter information including ($R_T, C_T$). This analysis will investigate the solution uniqueness of the Eqs. \ref{p1:eq1}-\ref{p1:wk}, which can be an informative prior with clinical relevance. In Fig. \ref{p1:2by4}(a), the model is used in a purely ``noninvasive'' configuration, taking only the easily-measured parameters $X = (\lambda_L, \lambda_D, \lambda_C, HR, SBP, DBP)$ as inputs and predicting $CO$, it yields strong correlation ($r \approx 0.94$) but with significant dispersion. A Bland-Altman analysis shows limits of agreement of approximately $\pm 1\,\text{L/min}$, which is large relative to the full physiological range of $CO$. This wide error margin indicates that, although the central parameters encode partial information about flow, the inverse problem remains effectively ill-posed: multiple combinations of the unobserved terminal parameters can reproduce nearly identical pressure signatures, and the network correspondingly struggles to converge to a unique solution across many virtual subjects.

Once either the extra pressure measurement site at the radial ($rDBP, rSBP$), or the terminal resistance and compliance ($R_T, C_T$) are supplied as inputs (Fig. \ref{p1:2by4} (b, c)), the inverse mapping becomes essentially identifiable. In this setting the model recovers $CO$ with near-perfect accuracy, producing mean errors within $\pm 0.03\,\text{L/min}$ and $\pm 0.01\,\text{L/min}$, respectively.

To investigate the minimal requirements for an optimal prediction, we will examine each parameter's predictive information content with respect to the solution of $CO$. Let \( X \in \mathcal{X} \subset \mathbb{R}^6\) denote the parameter vector which can be clinically inferred:

\begin{equation}
X = (\lambda_D,\lambda_L,\lambda_C,
  HR,DBP,SBP),
\end{equation}

\noindent and let $Y=CO$ be the true cardiac output computed by the
1-D hemodynamic solver. For this inverse scenario (Fig. \ref{p1:2by4}(a)), we consider a neural network surrogate \(f_\theta : X \to \mathbb{R}\) trained by minimizing the mean-squared error $\mathcal{L}(\theta)
= \mathbb{E}\big[(Y - f_\theta(X))^2\big]$. Now let $\widehat Y_{\mathrm{opt}} = f_\theta(X, \lambda_{C_T},\lambda_{R_T})$ denote the optimal predictor, obtained when all relevant inputs are provided (full model with Windkessel parameters, Fig. \ref{p1:2by4}(c)). Its irreducible variance $\sigma_{\mathrm{opt}}^{2}$ can be calculated from the coefficient of determination as follows:

\[
\sigma_{\mathrm{opt}}^{2} 
= \mathbb{E}\!\left[\operatorname{Var}(Y\mid X)\right] 
= \operatorname{Var}(Y - \widehat{Y}_{\mathrm{opt}}),
\]
\noindent where the total variance of the true $CO$ can be decomposed into the explained and unexplained (residual) components:
\[
\operatorname{Var}(Y) = \operatorname{Var}(\widehat{Y}_{\mathrm{opt}}) + \operatorname{Var}(Y - \widehat{Y}_{\mathrm{opt}}).
\]
\[
\frac{\operatorname{Var}(Y - \widehat{Y}_{\mathrm{opt}})}{\operatorname{Var}(Y)} = 1 - R_{\mathrm{opt}}^2 = 1 - 0.996 = 0.004 = 0.4\%.
\]

Thus the irreducible variance percentage for the optimal predictor is less than $1\%$, which makes it equivalent with the true $CO$. Now, let $\widehat Y_{\mathrm{red}}$ be a reduced predictor that excludes some inputs (e.g.\ omitting $R_T$). Its residual variance is:

\[
\overline{\sigma}_{\mathrm{red}}^{2} = \frac{\operatorname{Var}(Y - \widehat{Y}_{\mathrm{red}})}{\operatorname{Var}(Y)} = 1 - R_{\mathrm{red}}^2.
\]

The additional unexplained variance caused by the missing inputs is $\Delta \sigma_{\mathrm{}}^{2} = \sigma_{\mathrm{red}}^{2} - \sigma_{\mathrm{opt}}^{2}$, and in normalized form, 
\begin{equation}
\Delta \overline \sigma_{\mathrm{u}}^{2} = \frac{\Delta \sigma_{\mathrm{u}}^{2}}{\operatorname{Var}(Y)}
= 
  R_{\mathrm{opt}}^{2} - R_{\mathrm{red}}^{2}.
\end{equation}

This quantity measures precisely how much predictive information is lost by removing those inputs, or the irreducible uncertainty in $CO$ due to the missing variables. In Table $\ref{p1:t2}$ we consider all possible sub-cases that fall within the three main scenarios.

\begin{table}[h]
\centering
\caption[Inverse cases with the percentage of irreducible uncertainty]{Inverse cases with the percentage of irreducible uncertainty \textbf{$\Delta \overline \sigma_{\mathrm{u}}^{2}$}.}
\begin{tabular}{ccccc}
\hline
\textbf{Case} & \textbf{$r$} & \textbf{$R^2$} & \textbf{Error (\%)} & \textbf{$\Delta \overline \sigma_{\mathrm{u}}^{2}$} (\%)\\
\hline
$f_\theta(X)$ & 0.942 & 0.887 & 0.6 & 11\\
\hline
$f_\theta(X, rDBP, rSBP)$ & 0.984 & 0.968 & 0.32 & 2.7 \\
$f_\theta(X, rDBP)$ & 0.945 & 0.893 & 0.58 & 9.6 \\
$f_\theta(X, rSBP)$ & 0.982 & 0.964 & 0.3 & 3.1 \\
\hline
$f_\theta(X, \lambda_{R_T},\lambda_{C_T})$ & 0.998 & 0.996 & 0.1 & - \\
$f_\theta(X, \lambda_{R_T})$ & 0.998 & 0.996 & 0.12 & 0 \\
$f_\theta(X, \lambda_{C_T})$ & 0.994 & 0.989 & 0.19 & 0.6 \\

\hline
\label{p1:t2}
\end{tabular}
\end{table}
\vspace{-1em}
From a practical standpoint, a $11\%$ of information loss when only $X$ is taken into account is acceptable for a rough estimation of the $CO$, even if this can lead to errors of $20\%$ for certain patients, as shown in Fig. \ref{p1:2by4}(a). However, in clinical practice this error is expected to be more significant due to several complex physiological factors which are not accounted for by the 1-D numerical model. Interestingly, as also shown in Table $\ref{p1:t2}$, an additional pressure reading at the radial artery is enough to provide a very high correlation of $r=0.984$, with $2.7\%$ of unexplained variance, which is a strong indication that the accuracy in prediction $CO$ can be greatly enhanced via an additional pressure measurement site, even when the terminal Windkessel parameters are unknown.

As expected, using only $R_T$ still constrains the parameter space sufficiently to maintain strong correlation, with limits of agreement around $\pm 0.15\,\text{L/min}$ (Fig. \ref{p1:2by4} (c)). This reflects the dominant contribution of terminal resistance to the mean pressure-flow relationship 
and its central role in eliminating most of the ambiguity present in the reduced 
input set. A particularly interesting behavior emerges when only $C_T$ is provided (Table \ref{p1:t2}). Despite representing merely $1$--$2\%$ of the total arterial compliance and exerting only a 
minimal direct influence on central hemodynamics, its inclusion still yields an 
excellent correlation ($r = 0.996$) with limits of agreement of about 
$\pm 0.2\,\text{L/min}$, apart from a small number of outliers. This suggests that even 
weakly sensitive parameters can impart enough structural information to partially 
regularize the inverse problem, reducing the admissible solution manifold and guiding 
the network toward a nearly unique CO estimate.



\begin{figure}[H]
 \centering
 \includegraphics[width=0.9\textwidth]{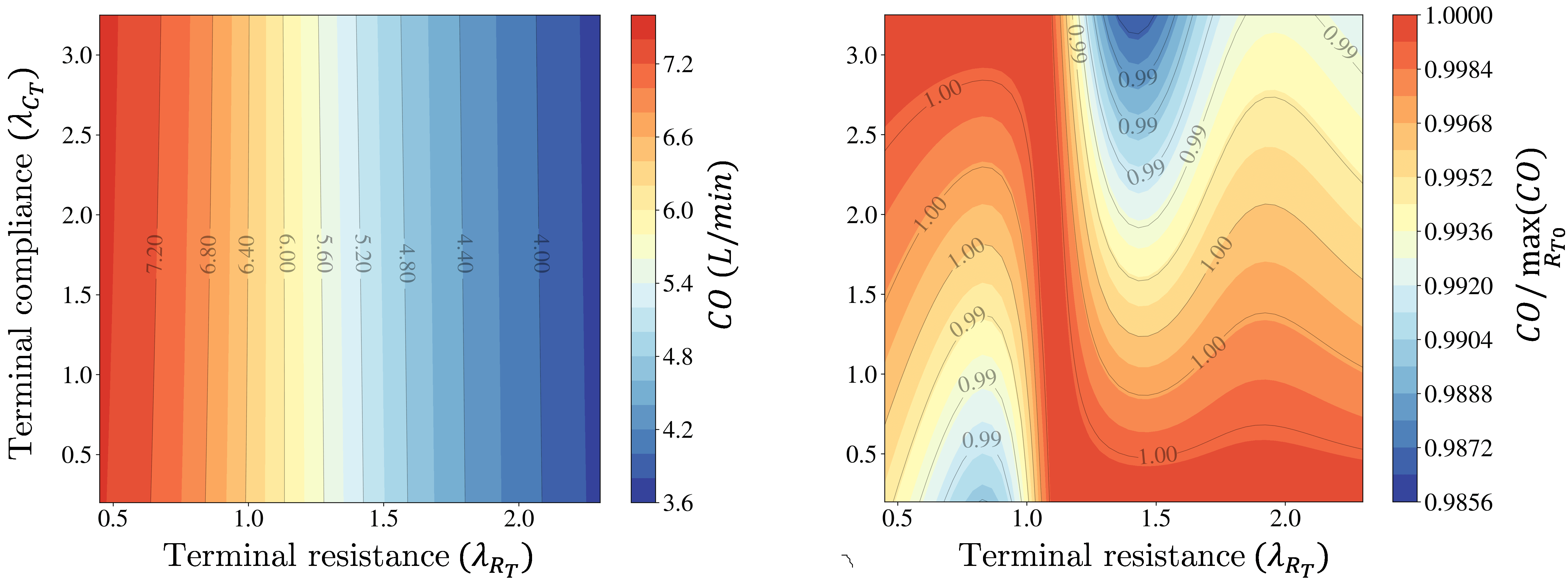}
 \caption[$R$-$C$ vs $CO$ grids]{ \textbf{$R$-$C$ vs $CO$ grids:} (a) Indicative patient contour show that the terminal resistance has a strong monotonic influence on $CO$. (b) To isolate the effect of $C_T$, we normalize each $R_T$ column with its max $CO$ value to reveal that there are at most 4 solutions of the same $C_T$, which constrains the possible combinations even when only $C_T$ is given as input along with $X$. }
 \label{p1:unique}
\end{figure}

To further investigate the cause of the unexplained variance due to $C_T$, we fix an input state $X_0$, which represents one of the patients with the worst correlation, and evaluate $f_{\theta}(X_0, R_T, C_T)=CO$ on a 2-D grid varying the terminal compliance and resistance multipliers (Fig. \ref{p1:unique}(a)). The dominant effect of $R_T$ is apparent from the monotonic increase for every fixed $C_{T0}$ value, so to isolate the sensitivity of $CO$ to $C_T$, for each fixed $R_{T0}$, we normalize with the maximum $CO$ value of that column (Fig. \ref{p1:unique}(b)). This removes the dominant effect of $R_T$ on the absolute scale of $CO$, leaving only the relative variation of $CO$ across $C_T$. Empirically, we observe that for any admissible normalized $CO$ level $\alpha$ and any fixed $C_T=C_{T0}$, the one-dimensional inverse problem $f(C_{T0},R_T) = \alpha$ has at most $m$ solutions in $R$, with $m\le 4$ over the sampled domain. Thus, although $(C_T,R_T)\mapsto \tilde f(C,R)$ is not injective, conditioning on $C_T$ reduces the two-dimensional inverse ambiguity for $(C_T,R_T)$ at a given output level $\alpha$ from a 2-D continuum (the full contour $(C_T,R_T)$-plane) to at most $m$ isolated points.

\subsection{Clinical application: aortic noninvasive hemodynamics}
While numerical simulations provide a powerful tool for hemodynamic analysis, it remains essential to quantify their deviation from real-world applications and assess the extend of their practical use. For that reason, we also employ the surrogate model $f_\theta(X)$ of Table \ref{p1:t2}, for the inverse estimation of both $CO$ and $cSBP$ in a clinical setting, which constitute two key indicators of cardiac function. The clinical dataset we choose to test on for this study is composed of 20 healthy patients of relatively young age, while the measurement protocol is described in detail on the original publication \cite{papaioannou2014first}.

\begin{figure}[H]
 \centering
 \includegraphics[width=0.7\textwidth]{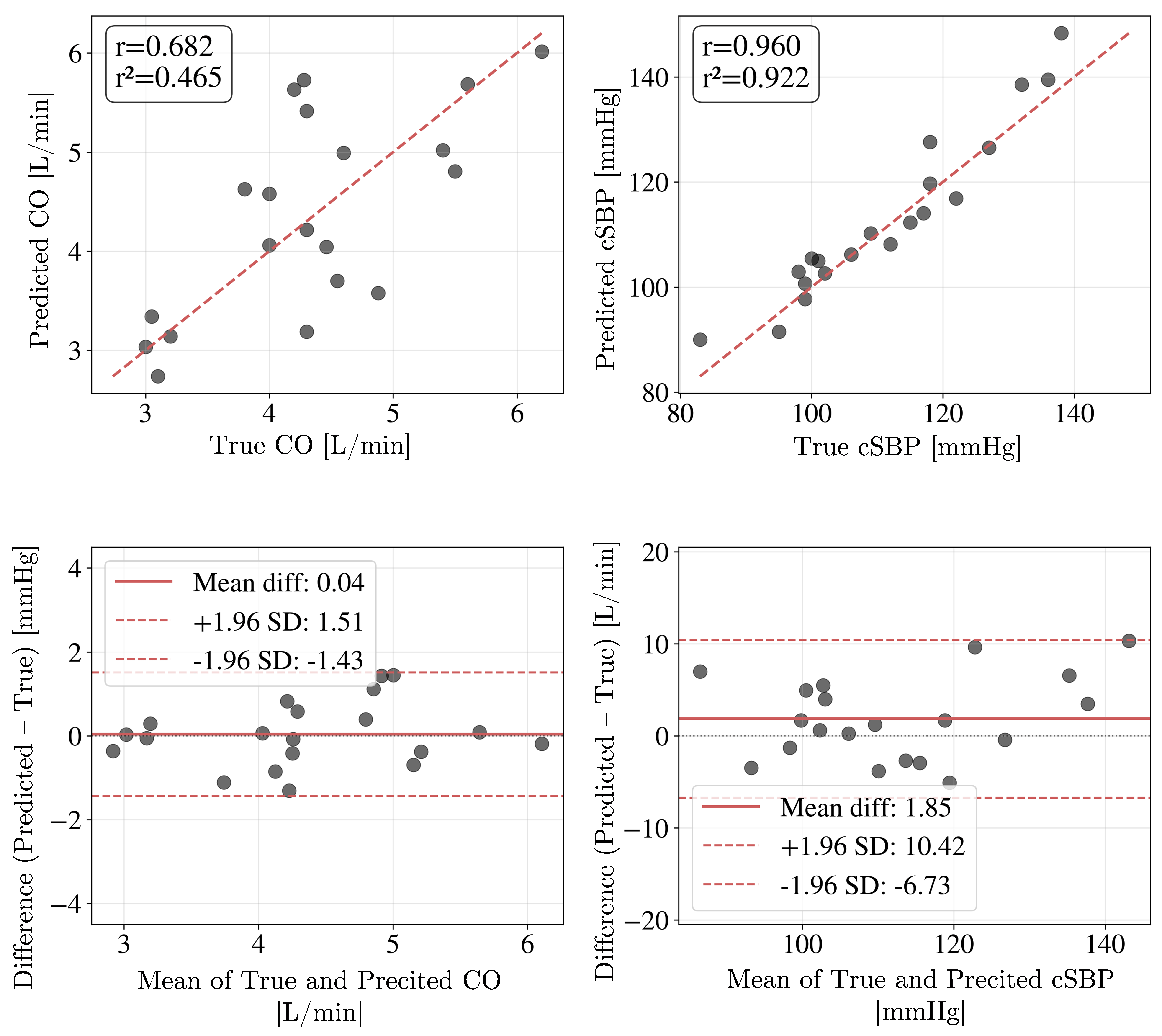}
 \caption[Cardiac hemodynamics prediction]{ \textbf{Cardiac hemodynamics prediction:} Using only noninvasive inputs $X$, the surrogate (a) captures the dominant flow across the clinical dataset, while (b) it provides a very accurate estimate for the aortic pressure. }
 \label{p1:clinical}
\end{figure}

For the training dataset, we use the clinically-matched virtual population of Section \ref{p1:matching}. The prediction results are shown in Fig. \ref{p1:clinical} for both output variables. Having seen only numerically derived patient solutions, our model attains correlations of $r=0.615$ for the $CO$, and $r=0.959$ for the $cSBP$, within errors of $\pm 1.6 L/min$ and $\pm 7.5 mmHg$, respectively. Although the model shows noticeable scatter for the $CO$ estimation, it is able to capture the dominant flow characteristics and especially the aortic blood pressure trends across the full range of our patients. It is worth noting that training on the entire \emph{in silico} population (including the non-admissible cases) yielded a $CO$ correlation of $r=0.5$, which further highlights the importance of our distribution-matching methodology workflow of Fig. \ref{p1:workflow}.

\newpage
\section{Limitations}
The current 1-D formulation reflects healthy physiology under idealized assumptions and does not capture the full variability seen across diseased states, age groups, or other patient-specific factors; these gaps would require targeted model extensions or the integration of complementary clinical information during training. The sensitivity matrix and hemodynamic parameters should be viewed as qualitative guides rather than definitive physiological relations, since true responses can deviate substantially across broader clinical populations. Available clinical datasets are typically sparse, cardiac output in particular is prone to large measurement errors, so they often support only population-level statistical use rather than precise individualized calibration. For example, the Asklepios dataset could not yield a correlation higher than $0.4$, with significant errors for the $CO$ (even when attempting to train surrogates using purely clinical data), suggesting the existence of additional unknown factors affecting the bloodflow or noise in the measurements. Ultimately, any clinical deployment demands cross-validation against multiple independent reliable datasets to demonstrate robustness, with the 1-D surrogate serving as a foundational layer for such expanded studies.

\section{Summary}


In this work, we develop an efficient methodology for training a surrogate model based on a physiologically grounded virtual cohort of 1-D arterial simulations, enabling real-time prediction, validation, and interpretation of hemodynamic behavior. By preserving the multivariate structure of the Asklepios clinical dataset, the virtual cohort provides realistic parameter correlations, while the neural surrogate reproduces arterial pressure and $CO$ across the admissible physiological range. The model performs instantaneous a priori rejection of non-physiological inputs, reducing the cost of generating targeted \emph{in silico} populations. It is also deployed for global sensitivity analysis and the construction of generalized pressure-response surfaces, revealing how dominant physiological parameters and their interactions shape systolic and diastolic arterial pressure.

The surrogate is also used to quantify the degree to which $CO$ is identifiable from patient-specific parameters alone, and to demonstrate how terminal resistance and compliance progressively constraint the parameter-output manifold. The resulting analysis shows the existence of non-uniqueness until the terminal parameters are supplied, with resistance having the strongest effect, while the compliance is able to provide an informative direction in the admissible solution space. On the practical side, since the Windkessel parameters are unknown, we show that providing an additional pressure reading on the radial artery (e.g. wrist) significantly reduces the uncertainty in predicting aortic hemodynamic parameters of $CO$ and $cSBP$. These insights provide a theoretical expectation on the optimal methodology for fast and accurate noninvasive monitoring of cardiovascular function during clinical scenarios.

Finally, we apply our surrogate on a clinical dataset, which yields satisfactory correlations for $CO$ and even better for $cSBP$. Attempts for testing on the larger dataset of Asklepios did not yield as good results, suggesting that mapping parameter-output manifolds or constructing generalized hemodynamic fields is not attainable using clinical data alone: measurements could be sparse, heterogeneous, and noisy, making numerical modeling indispensable for resolving the full multidimensional landscape. Clinical datasets are therefore useful to capture broad trends and correlation patterns rather than to provide exact patient-specific solutions. Future work will incorporate age-dependent markers to extend this framework to a wider population study. Along with additional clinical datasets, the surrogate can be a reliable foundation model, serving as an informative prior for specialized studies on disease indices and live health monitoring.

 \bibliographystyle{elsarticle-num} 
 \bibliography{cas-refs}

@article{reymond2009validation,
  title={Validation of a one-dimensional model of the systemic arterial tree},
  author={Reymond, Philippe and Merenda, Fabrice and Perren, Fabienne and Rufenacht, Daniel and Stergiopulos, Nikos},
  journal={American Journal of Physiology-Heart and Circulatory Physiology},
  volume={297},
  number={1},
  pages={H208--H222},
  year={2009},
  publisher={American Physiological Society}
}

@article{bikia2019noninvasive,
  title={Noninvasive cardiac output and central systolic pressure from cuff-pressure and pulse wave velocity},
  author={Bikia, Vasiliki and Pagoulatou, Stamatia and Trachet, Bram and Soulis, Dimitrios and Protogerou, Athanase D and Papaioannou, Theodore G and Stergiopulos, Nikolaos},
  journal={IEEE journal of biomedical and health informatics},
  volume={24},
  number={7},
  pages={1968--1981},
  year={2019},
  publisher={IEEE}
}

@article{bikia2020noninvasive,
  title={Noninvasive estimation of aortic hemodynamics and cardiac contractility using machine learning},
  author={Bikia, Vasiliki and Papaioannou, Theodore G and Pagoulatou, Stamatia and Rovas, Georgios and Oikonomou, Evangelos and Siasos, Gerasimos and Tousoulis, Dimitris and Stergiopulos, Nikolaos},
  journal={Scientific reports},
  volume={10},
  number={1},
  pages={15015},
  year={2020},
  publisher={Nature Publishing Group UK London}
}

@article{papaioannou2014first,
  title={First in vivo application and evaluation of a novel method for non-invasive estimation of cardiac output},
  author={Papaioannou, Theodore G and Soulis, Dimitrios and Vardoulis, Orestis and Protogerou, Athanase and Sfikakis, Petros P and Stergiopulos, Nikolaos and Stefanadis, Christodoulos},
  journal={Medical engineering \& physics},
  volume={36},
  number={10},
  pages={1352--1357},
  year={2014},
  publisher={Elsevier}
}

@article{bikia2021determination,
  title={Determination of aortic characteristic impedance and total arterial compliance from regional pulse wave velocities using machine learning: an in-silico study},
  author={Bikia, Vasiliki and Rovas, Georgios and Pagoulatou, Stamatia and Stergiopulos, Nikolaos},
  journal={Frontiers in bioengineering and biotechnology},
  volume={9},
  pages={649866},
  year={2021},
  publisher={Frontiers Media SA}
}

@article{bikia2021assessment,
  title={On the assessment of arterial compliance from carotid pressure waveform},
  author={Bikia, Vasiliki and Segers, Patrick and Rovas, Georgios and Pagoulatou, Stamatia and Stergiopulos, Nikolaos},
  journal={American Journal of Physiology-Heart and Circulatory Physiology},
  volume={321},
  number={2},
  pages={H424--H434},
  year={2021},
  publisher={American Physiological Society Rockville, MD}
}

@article{AsklepiosRietzschel2007,
    author = {Rietzschel, Ernst-R and De Buyzere, Marc L and Bekaert, Sofie and Segers, Patrick and De Bacquer, Dirk and Cooman, Luc and Van Damme, Piet and Cassiman, Peter and Langlois, Michel and van Oostveldt, Patrick and Verdonck, Pascal and De Backer, Guy and Gillebert, Thierry C and the Asklepios investigators},
    title = {Rationale, design, methods and baseline characteristics of the Asklepios Study},
    journal = {European journal of cardiovascular prevention and rehabilitation},
    volume = {14},
    number = {2},
    pages = {179-191},
    year = {2007},
    month = {04},
    issn = {1741-8267},
}

@article{AsklepiosSegers2007,
author = {Patrick Segers  and Ernst R. Rietzschel  and Marc L. De Buyzere  and Sebastian J. Vermeersch  and Dirk De Bacquer  and Luc M. Van Bortel  and Guy De Backer  and Thierry C. Gillebert  and Pascal R. Verdonck },
title = {Noninvasive (Input) Impedance, Pulse Wave Velocity, and Wave Reflection in Healthy Middle-Aged Men and Women},
journal = {Hypertension},
volume = {49},
number = {6},
pages = {1248-1255},
year = {2007},
}

@phdthesis{langewouters1982visco,
  title={Visco-elasticity of the human aorta in vitro in relation to pressure and age},
  author={Langewouters, Gerardus Johannes},
  year={1982},
  school={Krips Repro}
}

@article{segers2008three,
  title={Three-and four-element Windkessel models: assessment of their fitting performance in a large cohort of healthy middle-aged individuals},
  author={Segers, Patrick and Rietzschel, ER and De Buyzere, ML and Stergiopulos, N and Westerhof, N and Van Bortel, LM and Gillebert, Thierry and Verdonck, PR},
  journal={Proceedings of the Institution of Mechanical Engineers, Part H: Journal of Engineering in Medicine},
  volume={222},
  number={4},
  pages={417--428},
  year={2008},
  publisher={SAGE Publications Sage UK: London, England}
}

@article{lu2006continuous,
  title={Continuous cardiac output monitoring in humans by invasive and noninvasive peripheral blood pressure waveform analysis},
  author={Lu, Zhenwei and Mukkamala, Ramakrishna},
  journal={Journal of Applied Physiology},
  volume={101},
  number={2},
  pages={598--608},
  year={2006},
  publisher={American Physiological Society}
}

@article{wolak2008aortic,
  title={Aortic size assessment by noncontrast cardiac computed tomography: normal limits by age, gender, and body surface area},
  author={Wolak, Arik and Gransar, Heidi and Thomson, Louise EJ and Friedman, John D and Hachamovitch, Rory and Gutstein, Ariel and Shaw, Leslee J and Polk, Donna and Wong, Nathan D and Saouaf, Rola and others},
  journal={JACC: Cardiovascular Imaging},
  volume={1},
  number={2},
  pages={200--209},
  year={2008},
  publisher={American College of Cardiology Foundation Washington, DC}
}

@article{rietzschel2007rationale,
  title={Rationale, design, methods and baseline characteristics of the Asklepios Study},
  author={Rietzschel, Ernst-R and De Buyzere, Marc L and Bekaert, Sofie and Segers, Patrick and De Bacquer, Dirk and Cooman, Luc and Van Damme, Piet and Cassiman, Peter and Langlois, Michel and van Oostveldt, Patrick and others},
  journal={European Journal of Preventive Cardiology},
  volume={14},
  number={2},
  pages={179--191},
  year={2007},
  publisher={Oxford University Press}
}

@article{shi2011review,
  title   = {Review of {Z}ero-{D} and {1}-{D} Models of Blood Flow in the Cardiovascular System},
  author  = {Shi, Yubo and Lawford, Patricia and Hose, D. Rodney},
  journal = {BioMedical Engineering OnLine},
  year    = {2011},
  volume  = {10},
  pages   = {33},
}

@article{formaggia2003oned,
  title   = {One-dimensional models for blood flow in arteries},
  author  = {Formaggia, Luca and Lamponi, Daniele and Quarteroni, Alfio},
  journal = {Journal of Engineering Mathematics},
  year    = {2003},
  volume  = {47},
  number  = {3-4},
  pages   = {251--276},
}

@article{sherwin2003vascular,
  title   = {One-dimensional modelling of a vascular network in space-time variables},
  author  = {Sherwin, Spencer J. and Franke, Volker and Peir{\'o}, Joaquim and Parker, Kim H.},
  journal = {Journal of Engineering Mathematics},
  year    = {2003},
  volume  = {47},
  number  = {3-4},
  pages   = {217--250},
  }

@article{alastruey2009pattern,
  title   = {Analysing the pattern of pulse waves in arterial networks: A time-domain study},
  author  = {Alastruey, Jordi and Parker, Kim H. and Peir{\'o}, Joaquim and Sherwin, Spencer J.},
  journal = {Journal of Engineering Mathematics},
  year    = {2009},
  volume  = {64},
  pages   = {331--351},
  }

@article{matthys2007pulse,
  title   = {Pulse wave propagation in a model human arterial network: Assessment of 1-{D} numerical simulations against in vitro measurements},
  author  = {Matthys, Koen S. and Alastruey, Jordi and Peir{\'o}, Joaquim and Khir, Ashraf W. and Segers, Patrick and Verdonck, Pascal R. and Parker, Kim H. and Sherwin, Spencer J.},
  journal = {Journal of Biomechanics},
  year    = {2007},
  volume  = {40},
  number  = {15},
  pages   = {3476--3486},
  }

@article{laurent2006stiffness,
  title   = {Expert consensus document on arterial stiffness: Methodological issues and clinical applications},
  author  = {Laurent, St{\'e}phane and Cockcroft, John and Van Bortel, Luc and Boutouyrie, Pierre and Giannattasio, Carmine and Hayoz, Daniel and Pannier, Bruno and Vlachopoulos, Charalambos and Wilkinson, Ian and Struijker-Boudier, Harry and others},
  journal = {European Heart Journal},
  year    = {2006},
  volume  = {27},
  number  = {21},
  pages   = {2588--2605},
  }

@article{vanbortel2012cfpwv,
  title   = {Expert consensus document on the measurement of aortic stiffness in daily practice using carotid-femoral pulse wave velocity},
  author  = {Van Bortel, Luc M. and Laurent, St{\'e}phane and Boutouyrie, Pierre and Chowienczyk, Phil and Cruickshank, John K. and De Backer, Tine and Filipovsk{\'y}, Jan and Huybrechts, Stefaan and Mattace-Raso, Francesco U. S. and Protogerou, Athanase D. and others},
  journal = {Journal of Hypertension},
  year    = {2012},
  volume  = {30},
  number  = {3},
  pages   = {445--448},
  }

@article{rietzschel2007asklepios,
  title   = {Rationale, design, methods and baseline characteristics of the {A}sklepios Study},
  author  = {Rietzschel, Ernst R. and De Buyzere, Marc L. and Bekaert, Sofie and Segers, Patrick and De Bacquer, Dirk and Cooman, Luc and Van Damme, Piet and Cassiman, Peter and Langlois, Michel and Van Oostveldt, Patric and others},
  journal = {European Journal of Cardiovascular Prevention \& Rehabilitation},
  year    = {2007},
  volume  = {14},
  number  = {2},
  pages   = {179--191},
  }

@article{mckay1979lhs,
  title   = {A comparison of three methods for selecting values of input variables in the analysis of output from a computer code},
  author  = {McKay, M. D. and Beckman, R. J. and Conover, W. J.},
  journal = {Technometrics},
  year    = {1979},
  volume  = {21},
  number  = {2},
  pages   = {239--245},
  }

@article{pfaller2022automated,
  title   = {Automated generation of {0D} and {1D} reduced-order models of patient-specific blood flow},
  author  = {Pfaller, Martin R. and Pham, Jonathan and Verma, Aekaansh and Pegolotti, Luca and Wilson, Nathan M. and Parker, David W. and Yang, Wenbo and Marsden, Alison L.},
  journal = {International Journal for Numerical Methods in Biomedical Engineering},
  year    = {2022},
  volume  = {38},
  number  = {10},
  pages   = {e3639},
  }

@article{pegolotti2024gnn,
  title   = {Learning reduced-order models for cardiovascular simulations with graph neural networks},
  author  = {Pegolotti, Luca and Pfaller, Martin R. and Rubio, Natalia L. and Ding, Ke and Brugarolas Brufau, Rita and Darve, Eric and Marsden, Alison L.},
  journal = {Computers in Biology and Medicine},
  year    = {2024},
  volume  = {168},
  pages   = {107676},
  }


\end{document}